# SeqGenSQL - A Robust Sequence Generation Model for Structured Query Language

Ning Li
UC Berkeley

Bethany Keller
UC Berkeley

Mark Butler
UC Berkeley

Daniel Cer
UC Berkeley, Google Research

## Abstract

We explore using T5 (Raffel et al. (2019)) to directly translate natural language questions into SQL statements. General purpose natural language that interfaces to information stored within databases requires flexibly translating natural language questions into database queries. The best performing text-to-SQL systems approach this task by first converting questions into an intermediate logical form (LF) (Lyu et al. (2020)). While LFs provide a convenient intermediate representation and simplify query generation, they introduce an additional layer of complexity and annotation requirements. However, weakly supervised modeling that directly converts questions to SQL statements has proven more difficult without the scaffolding provided by LFs (Min et al. (2019)). We approach direct conversion of questions to SQL statements using T5 (Raffel et al. (2019)), a pre-trained textto-text generation model, modified to support pointer-generator style decoding (See et al. (2017)). We explore using question augmentation with table schema information and the use of automatically generated silver training data. The resulting model achieves 90.5% execution accuracy on the WikiSQL (Zhong et al. (2017)) test data set, a new state-of-the-art on weakly supervised SQL generation. The performance improvement is 6.6% absolute over the prior state-of-the-art (Min et al. (2019)) and approaches the performance of state-ofthe-art systems making use of LFs.

## 1 Introduction

A significant portion of the world's structured information is stored in relational databases. Accessing this information typically requires custom front-end interfaces or specialized knowledge in expressing the information being sought as SQL queries. In principle, natural language questions are a very human friendly interface to any knowledge repository. However, while there are many specialized domain specific dialog systems that allow users to interact with database backed systems, it has proven more difficult to construct general purpose models for the conversion of textual questions into database queries. The current state-of-the-art for text-to-SQL generation minimally makes use of specialized logical forms to aid in the translation of questions to databases queries Lyu et al. (2020), with diminished performance from systems that eschew such representations (Min et al. (2019); Wang et al. (2019); Agarwal et al. (2019)).

Our work focuses on using a pre-trained state of the art sequence generation model T5 with carefully designed question augmentation, data augmentation and architecture modification.

Beginning with a T5-small model and, with the addition of natural language questions and column headers as part of the input, we demonstrate how the addition of features in the input such as data types and data sampling lead to improved outputs. Test results show that data types and data sampling contain features that can be learned by the model.

Even with 56,355 training samples, it's still arguably not enough data for large natural language processing models. To further expand the data set, we train a reverse trainer model to get access to an additional 250K silver data training samples.

As an attempt to reduce hallucination and encourage extraction (e.g. column names from a natural language question), we implement a gated extraction layer which decides whether the model should extract (from the natural language question) or generate from the decoder.

There are two leader boards for the WikiSQL data set: weakly supervised (without using logical form during training) and supervised (with logical form during training). [1] On the supervised leader board, there are two results: those with execution guided inference and those without execution guided inference.

The previous state of the art weakly supervised model (HardEM, Min et al. (2019)) achieved 83.9% execution accuracy on the test data set. On the supervised model leader board, IE-SQL (Ma 2020) achieves 87.8% execution accuracy without execution guided inference on the test data set and 92.5% execution accuracy with execution guided inference.

## 2  Background

WikiSQL contains 56,355 training examples, 8,421 dev examples and 15,878 test examples. This data set includes natural language questions, table ids, annotated SQL logical form and data for each table.

```
{'phase': 1,
 'table_id': '1-1000181-1',
 'question': 'Tell me what the notes are for South Australia ',
 'sql': {'sel': 5, 'conds': [[3, 0, 'SOUTH AUSTRALIA']], 'agg': 0}}
```

| State/territory | Text/background c | Format |
|---|---|---|
| Australian Capital Territory | blue/white | Yaa·nna |
| New South Wales | black/yellow | aa·nn·aa |
| South Australia | black/white | Snnn·aaa |

| Current slogan | Current ser | Notes |
|---|---|---|
| ACT · CELEBRATION OF A CENTUR | YIL·00A | Slogan screenprinted on plate |
| NEW SOUTH WALES | BX·99·HI | No slogan on current series |
| SOUTH AUSTRALIA | S000·AZD | No slogan on current series |

Logical form is annotated by hand and includes an index for a select column, an index for an aggregation function, indices for a where condition column, and indices for where condition operators and values.

The output is a complete SQL statement:

```
SELECT function(Column)
FROM (Table ID)
 WHERE [Condition1, condition2, ...]
```

SQL statements in the WikiSQL data set have the following properties:

1. One select column
2. One select column aggregation function
3. One table for each question
4. Multiple "where" conditions

Most of the research done for this task uses a slot filling approach: predicting columns, functions and conditions separately then assembling the prediction into a final SQL statement. The work done by Hwang et al. (2019) showed that fine-tuning pre-trained models can provide significant improvement in many areas. In X-SQL (He et al. (2019)), a BERT/MT-DNN based fine-tuned model has shown significant improvement in both logical form prediction and execution accuracy
as compared to previous models. Another state of the art model (supervised), Hybrid Ranking Network (Lyu et al. (2020)), is also based on a BERT/RoBERTa pretrained model and achieved 92% execution accuracy using annotated logical form and execution guided inference - predictions returning empty results will be dropped and the next most probable prediction is chosen.

## 3  Methods

Historically, sequence generation models like Seq2Seq (Zhong et al. (2017)) which use a sequence generating decoder architecture to generate SQL statements, had challenges ensuring correct SQL syntax due to "hallucination" (generating data that does not actually exist) from deep learning models.

In this paper, we argue that with state of the art sequence generation models like T5 and with carefully designed question augmentation and data augmentation, a sequence generating model can overcome this "hallucination" challenge and generate SQL statements in one step with precise syntax and good execution accuracy.

T5 is also based on transformer (Vaswani et al. (2017)) architecture with additional layers as a decoder for sequence generation which treats all NLP tasks as sequence to sequence generation. It is pre-trained on a data-rich task before being fine-tuned on a downstream task. To establish a baseline, we used T5-small as a base model and followed a commonly adopted practice of combining the natural language question and table columns as input with SQL statements as output.

<bos>Natural question<sep> table-id<sep>col1<sep>col2 <sep>...coln<eos>

The labels are executable SQL statements[2], such as

---

[1] https://github.com/salesforce/WikiSQL

[2] All database object names are quoted according to SQL syntax.

```
SELECT [notes] FROM [1-1000181-1]
WHERE [state/territory] = 'south australia'
```

### 3.1 Question Augmentation

Besides column names, we believe that data types also contain information to be learned for sequence generation. For example, aggregation functions such as avg and sum can only be applied to numeric data columns. Based on this belief, we construct the new input to include data types.

```
<bos>Natural question<sep>table-id
<sep>col1<sep>type1<sep>col2
<sep>type2<sep>...coln<eos>
```

### 3.2 Reversed Trainer Model

We also experiment with a "reversed trainer" model flipping input and output for our pre-trained T5 model and fine tuning it (Artetxe et al. (2019)). This reversed trainer model takes a SQL statement as input and predicts a natural language question. We train the "reversed trainer" model for 20 epochs, then we randomly generate SQL statements from the tables provided, pass generated SQL statements through the reversed model and predict the natural language questions (silver data). Finally, we use this silver data to further fine tune the primary model. In our experiment, we generate 250,000 training samples in addition to the original 56,355 training samples.

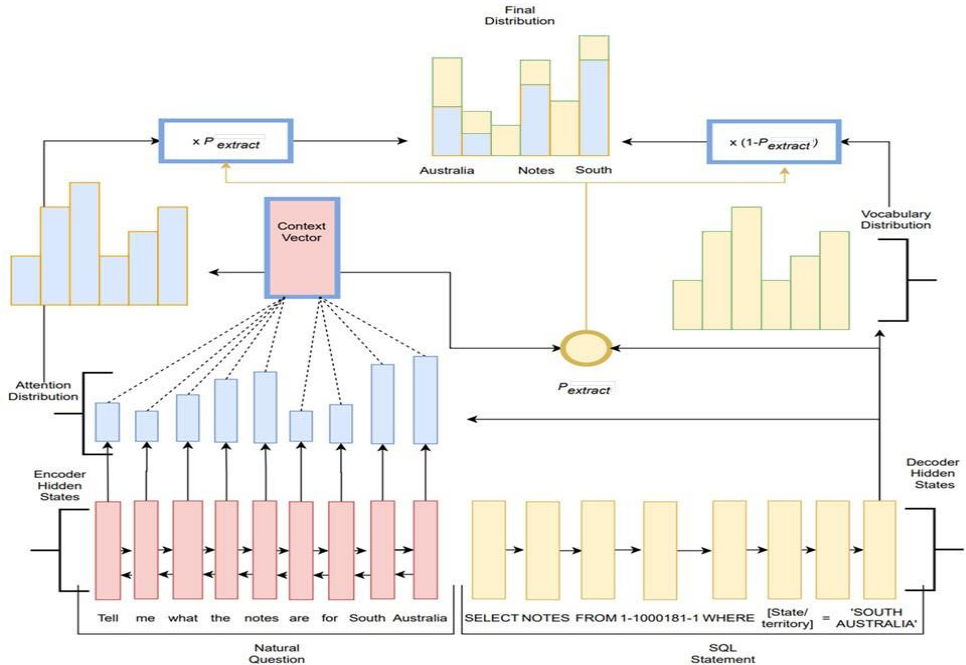

Figure 1: Gated Extraction Layer

Additionally, we borrow from the intuition of how humans write SQL statements - we examine the first few records to learn about the patterns and data format to guide us to write an informed statement. This technique further enhances the input to include the first few samples from each table. We experiment with sampling 1 row, sampling 2 rows and sampling 3 rows. Simply sampling 1 row showed significant improvement while sampling 2 or 3 produced better outcomes as shown in Table 1.

**Reversed Trainer Model**
trainerModel = Finetune T5 using ([SQL,Question]) pairs
while:
    selColIdx = random(len(table columns))
    aggFuncIdx = random(len(aggFuncList))
    numCondition = random(4)
    for cond in numCondition:
        condColIdx = random(len(table columns))
        condValueIdx = random(len(table rows))
    trainerSQL = Compose SQL statement using
            selColIdx, aggFuncIdx and condString
    trainerQuestion = trainerModel.generate(trainerSQL)

### 3.3 Gated Extraction Network

Like other sequence generation models (Zhong et al. (2017)), T5 also hallucinates and generates completely new words:

```
In which country is the city of Netanya?
Pred: select [country] from [1-14937957-1] where [city] = 'netheranya'
True: select [country] from [1-14937957-1] where [city] = 'netanya'
```

To reduce hallucination and encourage extraction, we implement a gated extraction T5 network. Similar to a Pointer Generation Network (See et al. (2017)), we implement a cross layer attention layer between the encoder($H_{enc}$) and decoder ($H_{dec}$). Then we create a gate layer from the attention layer to control whether the output shouldd be generated by the decoder or extracted from the encoder.

The cross layer attention layer is implemented the same way as the T5 cross attention layer where the score is the product of $H_{enc}$ and $H_{dec}$: $q = linear(H_{dec})$ $k = v = linear(H_{enc})$

$$Score = q * v$$

The context is then calculated using *Score*, $H_{dec}$ and *softmax* function:

$$Context = FF(softmax(Score) * v)$$

The final gate is a probability based on the hidden state of the decoder and *Context* wrapped inside a sigmoid function:

$$lnorm_{con} = LayerNorm(H_{dec})$$ $$lnorm_{Context} = LayerNorm(Context)$$

$$P_{ext} = sigmoid([lnorm_{con}, lnorm_{Context}])$$

The final step is to merge both extraction and generation together using element wise operation:

$$O_{final} = (1 - P_{ext}) * O_{gen} + P_{ext} * O_{ext}$$

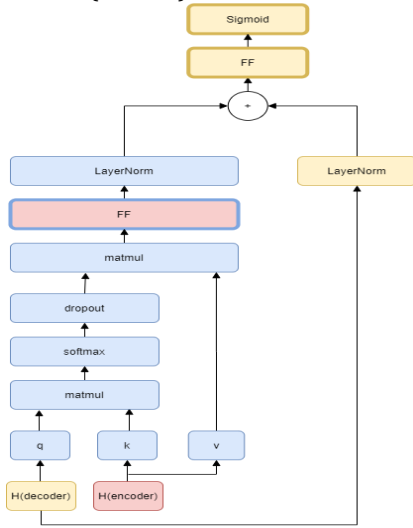

### 3.4 Execution Guided Inference

Even with slot filling models, it's challenging to generate perfect SQL prediction. To improve the prediction, Text2Sql (Chenglong Wang (2018)) introduced execution guided inference. Execution guided models send generated SQL statements to the SQL database engine and make adjustments if the database engine returns run-time errors or an empty result. While we are not convinced an empty result should be considered an error in practice, we did experiment with beam search using run-time error during execution.

To apply execution guided inference, we used beam search during inference to generate multiple output sequences for each question. Then, we sent each output sequence to the SQL database engine. If the SQL database engine returned a run-time error, we drop the current sequence and try the next output sequence.

```
Execution Guided Inference
predicts = Predict 3 SQL statements
finalOutput = []
for seq in predicts:
    try:
        run seq in SQL database engine
        finalOutput.append(seq)
        break
    finally:
        continue
```

The top performing models on the leader board applying execution guidance are getting significantly greater improvement in execution accuracy than SeqGenSQL. Error analysis shows SeqGenSQL only had only roughly 30 predictions using the dev data set with fixable execution errors. Since SeqGenSQL is already making such accurate predictions, execution guidance only showed small improvement.

## 4 Results

Using T5-small, our baseline model achieved 80.0% execution accuracy. With additional question augmentation like adding data types, data sampling and silver data, the test data execution accuracy improved to 88.5%, which exceeded the previous state of the art weakly supervised model (83.9%, Min et al. (2019)).

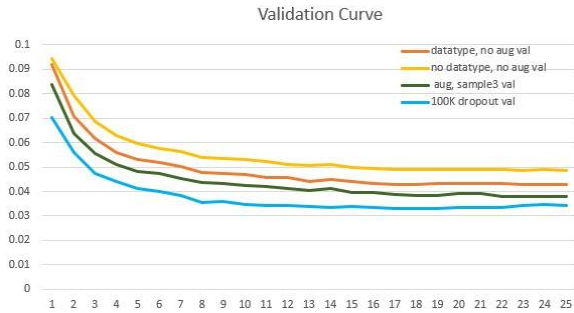

Using a reversed trainer model and input token dropout (randomly drop one masked token during training), we gained a further 1.6% improvement.

Our final model, built on a pre-trained T5-base (which contains 3x the parameters than T5-small) with data type and data sampling, trained on approximately 56,000 original training data samples

| Type | Category | Total Errors |
| --- | --- | --- |
| Invalid | Select Column | 16 |
|  | Agg Function | 0 |
|  | Where Column | 15 |
|  | Where Oper | 1 |
|  | Where Value | 94 |
| Wrong | Select Column | 138 |
|  | Agg Function | 396 |
|  | Where Column | 0 |
|  | Where Oper | 0 |
|  | Where Value | 0 |

As the list indicates, in a perfect world, execution guided inference could help improve the accuracy by 1.5% (all invalid errors are corrected by beam search).

| Model | Dev execution accuracy |
| --- | --- |
| T5 Baseline (T5-small) | 80.0 |
| +gated extraction | 79.1 |
| +gated extraction+datatype | 84.2 |
| +gated extraction+datatype+1sample | 85.6 |
| +gated extraction+datatype+2samples | 86.6 |
| +datatype+3samples | 86.8 |
| +gated extraction+datatype+3samples | 87.9 |
| +gated extraction+datatype+3samples+50K silver data | 88.5 |

Table 1: SeqGenSQL Results

and 250,000 silver training data samples, achieves 90.5% execution accuracy on the dev data set and 90.2% execution accuracy on the test data set. This model exceeds all models on the weakly supervised leader board and all models without execution guidance on the supervised leader board. SeqGenSQL lands 4th place overall on the combined leader board.[3]

## 5 Error Analysis and Discussion

Sequence generation simplifies the SQL statement generation process and achieves a one step prediction for this task with high accuracy.

Error analysis on the dev data set showed the most mistakes are on predicting the wrong aggregation functions. In the below table, "Invalid" error type means the predicted words cannot be found in a table column (if it's a column name) or cannot be found in the original question (if it's a where condition value). "Wrong" error type indicates the total number of predictions that don't match the gold label:

We further break down the errors into 5 classes: hallucination, multilingual text, ambiguous questions, missing information and data issues.

### 5.1 Hallucination

Even with a gated extraction network, hallucination still happens. In the following example, SeqGenSQL generated a new value "1962-001a" in the where condition instead of extracting the word "1962-011a" from the question:

> Question:
> How many alt names does 1964-011a have?
> Prediction: select count([alt name]) from [1-12141496-1] where [id] = '1962-001a'

### 5.2 Multilingual Text

WikiSQL is a crowd sourcing data set and contains multiple languages. T5 was trained on only English data. Therefore, the T5 tokenizer cannot interpret non-English characters:

---

[3] https://github.com/salesforce/WikiSQL

How many people live in 铅山县?

## 5.3 Ambiguity in Questions

Some questions contain words that are ambiguous. For example, the word "total" in the following question can be interpreted as a sum function in SQL:

> Question:
> What is the total brup for the team?
> Gold Label:
> select [brup] from [1-18064020-21] where [name] = 'total'

## 5.4 Missing Information

Some of the questions simply didn't provide sufficient information to compose an accurate SQL statement. In this example, the "gold label" (human-generated label for comparison) uses the "hardcover" column while SeqGenSQL chose to use "paperback". The difference between these two columns cannot be identified from the question.

> Question:
> How many publishers put out isbn 193700788x?
> Gold Label:
> select count([publisher]) from [1-16907214-1] where [hardcover]='isbn 193700788x'
> Prediction
> select count([publisher]) from [1-16907214-1] where [paperback] = 'isbn 193700788x'

## 5.5 Data Issues

There are cases where gold labels don't match the questions. In some of these cases, SeqGenSQL predicted correctly in manual verification, even though they don't match gold labels:

> Question:
> Which places have points larger than 10? Gold Label:
> select min([points]) from [2-10301911-6] where [place] > '10' Prediction select [place] from [2-10301911-6] where [points] > '10'

## 6 Future Research

Even though some of these errors (data issues), might not be solved with a better model, we still consider the following things that could be solved by improving the model. For example:

1. Larger base model

There are 5 pre-trained T5 models available. We experimented with T5-small and T5-base and observed improvement in accuracy using larger model. We assume larger base models could provide even more improvement.

| Model | Dev execution accuracy | Test execution accuracy |
| --- | --- | --- |
| Supervised | | |
| IE-SQL + EG (Ma 2020) | 92.6 | 92.5 |
| HydraNet + EG (Lyu et al., 2020) | 92.4 | 92.2 |
| X-SQL + EG (He et al., 2019) | 92.3 | 91.8 |
| EG Guided Decoding with BERT-Base Uncased (Guo and Gao, 2019) | 91.1 | 90.1 |
| HydraNet (Lyu et al, 2020) | 89.1 | 89.2 |
| X-SQL (He et al., 2019) | 89.5 | 88.7 |
| Weakly supervised | | |
| HardEM (Min et al., 2019) | 84.4 | 83.9 |
| SeqGenSQL(T5-base + 250K silver data) | 90.6 | 90.3 |
| SeqGenSQL + EG | 90.8 | 90.5 |

Table 2: WikiSQL Leader board

2. Improve the gated extraction layer

As we observed in error analysis, there is still a small amount of hallucination happening. Further investigation and adjustment could be done to improve execution accuracy in this area.

3. Include more sampling data Our tests showed more sampling data can improve accuracy. In most of our tests we used 3 samples in input. We could experiment further with a higher number of samples and observe whether it does/does not improve the model.

## 7 Conclusion

In the past, sequence generation hasn't been a popular choice for structured language generation due to syntactic challenges. Our paper shows that with a strong pre-trained cutting edge sequence generation model and carefully designed question augmentation, data augmentation and model modification, sequence generation can achieve great results even without execution guided inference and slot filling techniques. Our sequence generation model is able to predict SQL statements in one step with 90.5% execution accuracy.